\documentclass[unnumsec,webpdf,contemporary,large]{oup-authoring-template}%
\graphicspath{{Fig/}}

\usepackage[mathlines, switch]{lineno}
\usepackage{subcaption}
\usepackage{arydshln}

\usepackage{changes}
\setaddedmarkup{{\color{black}#1}}

\theoremstyle{thmstyleone}%
\theoremstyle{thmstyletwo}%
\theoremstyle{thmstylethree}%
\newtheorem{definition}{Definition}

\begin{document}

\journaltitle{Journal Title Here}
\DOI{DOI added during production}
\copyrightyear{YEAR}
\pubyear{YEAR}
\vol{XX}
\issue{x}
\access{Published: Date added during production}
\appnotes{Paper}

\firstpage{1}

%\subtitle{Subject Section}

\title[Short Article Title]{AutoPCR: Automated Phenotype Concept Recognition by Prompting}

\author[1]{Yicheng Tao}
\author[2]{Yuanhao Huang}
\author[2]{Yiqun Wang}
\author[2]{Xin Luo}
\author[1,2,$\ast$]{Jie Liu}

\address[1]{\orgdiv{Department of Electrical Engineering and Computer Science}, \orgname{University of Michigan}, \orgaddress{\street{2260 Hayward Street}, \postcode{48109}, \state{Michigan}, \country{United States}}}
\address[2]{\orgdiv{Department of Computational Medicine and Bioinformatics}, \orgname{University of Michigan}, \orgaddress{\street{100 Washtenaw Avenue}, \postcode{48109}, \state{Michigan}, \country{United States}}}

\corresp[$\ast$]{Corresponding author. \href{email:email-id.com}{drjieliu@umich.edu}}

\received{Date}{0}{Year}
\revised{Date}{0}{Year}
\accepted{Date}{0}{Year}

%\editor{Associate Editor: Name}

\abstract{
\textbf{Motivation:} Phenotype concept recognition (CR) is a fundamental task in biomedical text mining. However, existing methods either require ontology-specific training, making them struggle to generalize across diverse text styles and evolving biomedical terminology, or depend on general-purpose large language models (LLMs) that lack necessary domain knowledge.\\
\textbf{Results:} To address these limitations, we propose AutoPCR, a prompt-based phenotype CR method designed to automatically generalize to new ontologies and unseen data without ontology-specific training. To further boost performance, we also introduce an optional self-supervised training strategy.
Experiments show that AutoPCR achieves the best average and most robust performance across datasets. Further ablation and transfer studies demonstrate its inductive capability and generalizability to new ontologies.\\
\textbf{Availability and Implementation:} Our code is available at \url{https://github.com/yctao7/AutoPCR}.\\
\textbf{Contact:} \href{name@email.com}{drjieliu@umich.edu}\\
% , supporting applications such as clinical diagnostics and knowledge graph construction
% %AutoPCR performs CR in three stages: entity extraction using a hybrid of BioNER and syntax-based strategies, candidate retrieval via SapBERT, and entity linking through prompting an LLM. 
% \textbf{Supplementary information:} Supplementary data are available at \textit{Journal Name}
% online.
}

% \abstract{Check the journal's online author guidelines for guidance on requirements for the abstract and other components of the submitted manuscript. Abstracts must be able to stand alone and so cannot contain citations to the paper's references, equations, etc. Que cum aut etum qui ium dolupta ssequia autati odis demporepe ad et es alit rem repudaerae min et volorum re volupta nobit volectur aut fuga.} 

% \keywords{concept recognition, large language model}

% \keywords[Abbreviations]{CR, LLM}

% \otherabstract[Additional Abstract]{Use this element for elements such as Graphical abstract, Lay summary, Translated abstract etc. Que cum aut etum qui ium dolupta ssequia autati odis demporepe ad et es alit rem repudaerae min et volorum re volupta nobit volectur aut fuga.}

% \otherabstract[Graphical Abstract]{\colorbox{black!20}{\hbox to 0.97\textwidth{\vbox to 50pt{}}}}

% \boxedtext{Key Messages}{
% \begin{itemize}
% \item Key boxed text here.
% \item Key boxed text here.
% \item Key boxed text here.
% \end{itemize}}

\maketitle

%\begin{epigraph}
%Epigraph text. Ximporem qui reperov idempedit modio. Bisto imagnatem quae aceptis
%nobitae quid eum rae adignis quias-sit vellacc uptatur sunt quis rentis eaquasit alia deliquam
%rec-to consed unt. Empor sum ratur ressimusdae. Nam fugiae.
%\source{Epigraph source}
%\end{epigraph}

\section{Introduction}

Biomedical text mining plays a key role in unlocking clinical and scientific knowledge from unstructured data sources such as clinical notes and research articles. A fundamental step in this process is ontology-based concept recognition (CR), which aims to identify textual mentions of concepts defined in a given ontology from input text in a zero-shot setting. An ontology is a formal, structured representation of domain-specific knowledge curated by experts, consisting of standardized concepts with associated names, definitions, synonyms, and hierarchical relationships.
Phenotype CR, a specific instance of ontology-based CR using the Human Phenotype Ontology (HPO) \citep{gargano2024mode}, has become a central research focus. This is largely due to the availability of richly annotated datasets \citep{weissenbacher2023phenoid, lobo2017identifying, anazi2017expanding}, yet the task remains challenging because of the specialized and rapidly evolving nature of biomedical terminology and ontologies.
Phenotype CR plays a critical role in downstream biomedical applications. For example, genetic disease diagnostics require accurate identification of phenotype concepts in clinical notes \citep{labbe2023chatgpt}, and biomedical knowledge graph construction relies on robust CR from scientific literature to support integrative data analysis and knowledge discovery \citep{huang2024building}.
% \citep{soman2024biomedical}.

Traditionally, CR can be divided into two stages: (1) entity extraction, which identifies text spans to be linked to ontology concepts; and (2) entity linking, which maps these spans to semantically corresponding ontology concepts.
Early CR methods are primarily dictionary-based, which rely on lookup tables and string-matching techniques to identify concept mentions. While these approaches offer high precision, they suffer from low recall due to limited vocabulary coverage and inability to handle linguistic variations \citep{jonquet2009ncbo, taboada2014automated}. 
In recent years, researchers have increasingly turned to neural methods that fine-tune domain-specific pre-trained language models (PLMs), such as BioBERT \citep{lee2020biobert}, on the rich synonym sets provided by the ontology. 
By distilling knowledge from synonymous relationships, these models effectively mitigate the challenges of lexical heterogeneity.
However, as they are trained against a fixed ontology, re-training is required to recognize new concepts, limiting their usability for frequently updated ontologies like HPO as well as generalizability to unseen ontologies.
In contrast, large language models (LLMs), such as GPT-4 \citep{achiam2023gpt}, have demonstrated strong zero-shot learning capabilities, offering new possibilities for phenotype CR. Recent studies have shown that LLMs can effectively extract clinical information, where phenotype concepts abound, without domain-specific fine-tuning \citep{meoni2023large}. Despite their potential, \added{LLMs acquire ontology knowledge imperfectly during pretraining and cannot reliably disambiguate among the large number of highly specialized concepts in HPO without explicit ontology access \citep{labbe2023chatgpt, groza2024evaluation}.} To address these challenges, retrieval-augmented generation \cite[RAG]{lewis2020retrieval} has emerged as an effective technique to improve LLM's CR performance by incorporating relevant information retrieved from the ontology through semantic similarity-based retrieval mechanisms. However, \added{the existing RAG-based method \citep{shlyk2024real}} relies on general-purpose components that fail to capture the nuanced semantics of highly specialized and dynamic biological concepts.

In this study, we propose \textbf{AutoPCR}, an automated phenotype CR framework based on prompting, which consists of three sequential steps. 
\added{First, AutoPCR employs a unified entity extraction approach including BioNER and syntax-based strategies with coordinated phrase decomposition, to ensure that extracted entities are biologically meaningful with broad coverage across both free-form clinical notes and standardized biomedical abstracts.}
In the second step, AutoPCR retrieves candidate concepts using a domain-specific semantic similarity model. Although phenotype CR has been extensively studied, prior work has largely overlooked advances in distantly-supervised biomedical entity linking. AutoPCR adopts one of these models named SapBERT \citep{liu-etal-2021-self} to accurately identify semantically relevant concepts. \added{AutoPCR then applies confidence-threshold routing that directly accepts high-confidence matches while delegating ambiguous cases to further LLM-based linking, improving both precision and recall.}
In the final step, AutoPCR performs entity linking by prompting an LLM. Each entity and its associated candidate set are encoded into a structured prompt that includes the synonyms and definition of each candidate concept. This approach enables accurate disambiguation even without ontology-specific fine-tuning. To further enhance its capability in distinguishing similar concepts, we introduce \added{an optional} self-supervised fine-tuning strategy that \added{automatically constructs hard positive and negative training examples from ontology structure via SapBERT retrieval, enabling efficient ontology-specific adaptation without manual annotation.}
\added{Together, AutoPCR introduces three technical novelties: a unified entity extraction approach that combines BioNER and syntax-based strategies, a domain-specific retrieval pipeline with confidence-threshold routing, and a self-supervised strategy for annotation-free ontology adaptation.}

Experiments demonstrate the following advantages of AutoPCR: 
(1) \added{\textbf{Best average and robust performance} across various benchmarks}, changing the status \added{quo} that prompt-based methods are lagging behind dictionary-based and neural methods. 
(2) \textbf{Inductive capability}. AutoPCR maintains \added{strong} performance even without prior exposure to HPO, making it adaptable to the frequent updates of HPO. 
(3) \textbf{Generalizability}. AutoPCR transfers well to a new ontology without reconfiguration and supports rapid deployment within minutes, offering great potential for broad application.

% These components collectively enable AutoPCR's ``auto'' capability, as SapBERT offers broad biomedical concept coverage through training on UMLS, and LLMs provide generalizable reasoning for entity linking without ontology-specific adaptation. It is thus well-suited for deployment in low-resource or rapidly evolving biomedical domains. Our key contributions are:

% num of experiments, conclusion, empirical results support, no techinical details, why important (fill a gap, why other methods cannot achieve, big picture (use what to improve performance)

% \begin{itemize}[leftmargin=*, itemsep=0.0em]

% We introduce a dual-index ontology representation that leverages both definition-based and synonym-based retrieval mechanisms, combined with a reranking strategy to refine candidate selection.

\section{Related work}

\paragraph{Phenotype concept recognition}

Phenotype CR has grown into multiple methodological paradigms, including dictionary-based, neural, and prompt-based methods.

Dictionary-based methods use lookup tables or string similarity measures to identify entities by exhaustively matching input spans against ontology concepts. Tools such as NCBO \citep{jonquet2009ncbo}, OBO \citep{taboada2014automated}, Doc2HPO \citep{liu2019doc2hpo}, and ClinPhen \citep{deisseroth2019clinphen} exemplify this strategy. Recently, FastHPOCR \citep{groza2024fasthpocr} utilizes groups of morphologically equivalent words generated by GPT-4 to address lexical variations.
%These methods achieve high precision due to strict matching mechanisms but suffer from low recall, particularly in handling complex linguistic variations and unseen terms.
% Additionally, their limited ability to capture contextual semantics restricts their effectiveness in dynamic biomedical corpora.

Neural methods leverage deep learning architectures, such as convolutional neural networks (CNNs) and BERT \citep{devlin2019bert}, with supervised learning on ontology-derived synonym pairs to improve performance.
There are two common entity extraction strategies adopted by neural methods, serving as the foundation for subsequent entity linking. One \added{category} of methods, e.g., PhenoTagger \citep{luo2021phenotagger} and PhenoTagger++ \citep{qi2024improved}, adopts a syntax-based strategy that exhaustively generates entity spans after removing punctuation and function words. PhenoTagger predicts concepts from extracted entities by fine-tuning a BioBERT. 
PhenoTagger++ improves PhenoTagger by augmenting training data with LLM-generated synonyms and character edition, as well as introducing graph-based concept embeddings from the ontology structure. 
The other \added{category} of methods \added{uses} off-the-shelf BioNER tools such as Stanza \citep{zhang2021biomedical} to identify biomedical segments and then exhaustively extract entities from them. 
NCR \citep{arbabi2019identifying} encodes entities using pre-trained text embeddings followed by a CNN and matches them with learnable embeddings of ontology concepts after hierarchical aggregation. 
PhenoBERT \citep{feng2022phenobert} builds upon NCR with multiple CNNs for phenotype subcategories to retrieve candidate concepts and a trainable BioBERT to score each entity-concept pair.
Recently, PhenoBCBERT and PhenoGPT \citep{yang2024enhancing} train Bio+ClinicalBERT \citep{alsentzer2019publicly} and GPT-3 \citep{brown2020language} on manually labeled data to improve rare disease recognition.

% PhenoTagger \citep{luo2021phenotagger} combines dictionary-based methods with a pre-trained BioBERT \citep{lee2020biobert} model. 
% PhenoTagger++ \citep{qi2024improved} further integrates concept embeddings built from TransR \citep{lin2015learning} according to the ontology structure. 

% These approaches improve recall by leveraging the semantic understanding capabilities of LLMs. 
% However, they typically rely on task-specific training aligned with a fixed ontology. 
% When the ontology evolves rapidly--e.g., with monthly updates like HPO--these models require frequent retraining to recognize newly introduced concepts. 
% This increases maintenance and computational costs, making it less practical for fast-evolving domains.

LLMs, such as GPT-4, enable phenotype CR without the need of ontology-specific training, allowing for rapid adaptation to evolving biomedical vocabularies. Prompt-based methods, such as \citet{labbe2023chatgpt} and \citet{groza2024evaluation}, \added{employ} prompt engineering to directly extract biomedical concepts from text. REAL \citep{shlyk2024real} in a RAG style relies on LLMs to extract biomedical entities and perform entity linking. Candidate concepts are retrieved for each entity using a general-purpose embedding model based on names and definitions, and then passed to an LLM for final linking.
% Despite their versatility, prompt-based methods face challenges of high computational costs and low throughput, and typically underperform compared to fine-tuned approaches tailored to the target ontology.
% he advent of general-purpose LLMs, such as GPT-4, has enabled phenotype CR without the need for ontology-specific training, enabling rapid adaptation to evolving biomedical vocabularies and flexibility across different biomedical domains.

\paragraph{Distantly-supervised biomedical entity linking}

Distant supervision has emerged as a practical solution for biomedical entity linking, where manual annotation is costly and often infeasible. Given the Unified Medical Language System \cite[UMLS]{bodenreider2004unified}, a comprehensive biomedical ontology with over 4M concepts covering HPO, these methods leverage ontology-derived supervision, enabling large-scale training without labeled corpora. BioSyn \citep{sung-etal-2020-biomedical} aligns entities with concept synonyms using a combination of character-level features and learnable embeddings. SapBERT fine-tunes a PubMedBERT \citep{gu2021domain} through contrastive learning on hard concept name-synonym pairs. KrissBERT \citep{zhang-etal-2022-knowledge} incorporates concept context from PubMed \cite{canese2013pubmed} into contrastive learning and employs a cross-attention encoder to re-rank candidate concepts. Its inference strategy retains multiple context embeddings per concept, enabling context-aware matching. Recent work \citep{sasse2024disease} shows that synthetic entities generated by LLMs can further enhance linking performance, especially under distribution shifts, highlighting the complementary role of generative models in distant supervision.

% AutoPCR advances the RAG paradigm by incorporating a dual-index ontology representation that leverages both definition-based and synonym-based retrieval mechanisms. Its iterative LLM-prompted NER strategy enhances recall while maintaining precision, and its structure-aware entity linking approach integrates ontology hierarchies to improve concept alignment. By addressing the limitations of prior methods—such as the low recall of dictionary-based systems, data dependency of neural models, and the inflexibility of traditional linking strategies—AutoPCR offers a robust, ontology-guided solution for biomedical concept recognition.

\section{Method}

\begin{figure*}[!t]
    \centering
    \includegraphics[width=0.76\textwidth]{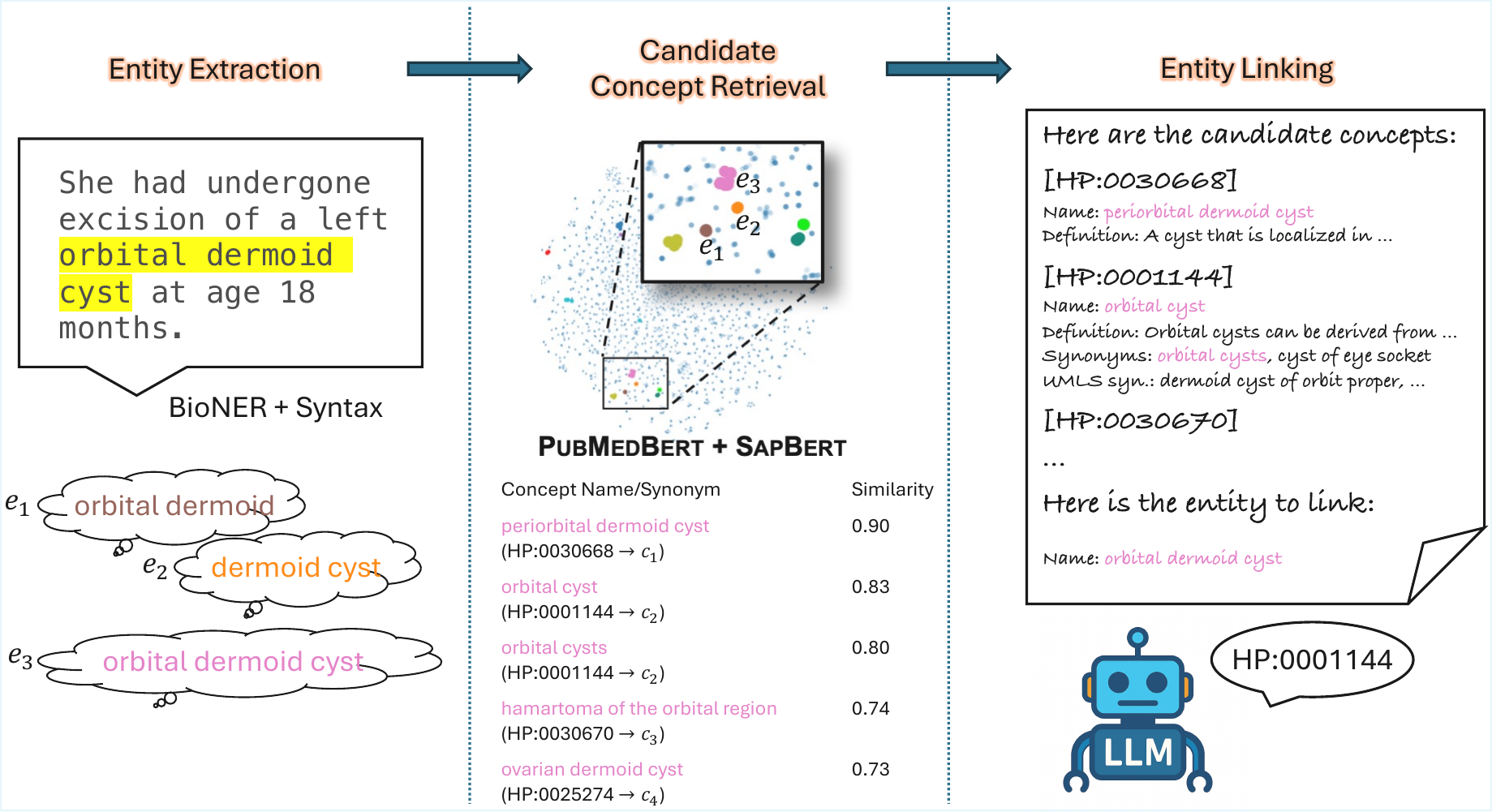}
    % For example, three entities. details. larger fonts. function name in title. 
    % The detailed prompt template is shown in Section~\ref{sec:el}.
    \caption{Architecture of AutoPCR. It performs concept recognition in three stages: entity extraction using BioNER and syntactic entity extraction (e.g., extracted entities $e_1$, $e_2$, and $e_3$), candidate concept retrieval via SapBERT initialized from PubMedBERT (e.g., retrieved concepts \added{$c_1$, $c_2$, $c_3$, and $c_4$} for $e_3$), and entity linking through prompting an LLM (e.g., linked concept \added{$c_2$}).}
    \label{fig:arch}
\end{figure*}

In this section, we present AutoPCR, an automated method for phenotype CR based on prompting. We begin with a formal definition of the CR task and describe how it can be decomposed into sequential subtasks. We then define each subtask rigorously and elaborate on how AutoPCR addresses them through an integrated and modular design. The overall architecture of AutoPCR is shown in Figure~\ref{fig:arch}.

% AutoPCR is designed to flexibly adapt to varying input styles and supports ontology-level supervision without requiring task-specific annotations. An overview of the problem formulation is provided first, followed by detailed descriptions of each module.

% \subsection{Problem formulation}

\begin{definition}[Concept recognition] Given an ontology $O{=}(C, I)$ with concepts $C{=}\{c_1, \dots, c_n\}$ and concept-level information $I$ (e.g., definitions and synonyms), for a piece of input text $T$, the task is to find $f$ such that
\begin{equation}
    f(O, T) = \{(i, j, c) \mid i, j \in [0, |T|],\; T[i\mathord{:}j] \sim c\in C\}
\end{equation}
which extracts entities from the input text with start and end offsets that can be mapped to semantically similar concepts in the ontology. 
\label{def:cr}
\end{definition}

We decompose the CR task $f$ into three sequential subtasks $f_{\text{EL}}\circ f_{\text{CCR}}\circ f_{\text{EE}}$, which are entity extraction $f_{\text{EE}}$, candidate concept retrieval $f_{\text{CCR}}$, and entity linking $f_{\text{EL}}$.

% \subsection{Entity extraction}
% \label{sec:eg}

\begin{definition}[Entity extraction] Given an ontology $O{=}(C, I)$ and a piece of input text $T$, the task is to find $f_{\text{EE}}$ such that
\begin{equation}
    f_{\text{EE}}(O, T) \supset \{(i, j) \mid i, j \in [0, |T|],\; T[i\mathord{:}j] \sim \exists c\in C\}
\end{equation}
\label{def:eg}
which extracts entity spans from the text that may correspond to ontology concepts.
\end{definition}

% We adopt two complementary strategies for extracting entities, tailored to different types of input text. For shorter, free-form text such as clinical notes, we follow a rule-based strategy inspired by PhenoTagger. The input text is split into sentences, tokenized, POS-tagged using NLTK \citep{bird2009natural}, and converted to lowercase. All $n$-gram spans ($n {\in} [2, 10]$) are then enumerated as candidate entities, excluding unigrams due to their limited variability and tendency to be misclassified as false positives. A part-of-speech filter is applied to eliminate spans that begin or end with punctuation or function words, including prepositions, conjunctions, and determiners.

We employ a unified entity extraction approach that combines BioNER and syntax-based strategies, making it applicable to both free-form clinical notes and standardized biomedical abstracts. For BioNER, we follow PhenoBERT and process each sentence using Stanza's BioNER model to extract biomedically relevant segments. To improve coverage, additional segments are generated by splitting on punctuation and conjunctions, and all resulting segments are used as windows for $n$-gram entity extraction ($n{\in}[2,10]$).
To better cover free-form text, we complement BioNER with a \emph{syntactic entity extraction} component, which identifies structurally coherent phrases through noun phrase extraction, coordinated phrase decomposition, and abbreviation recovery.
For noun phrase extraction, we employ the Berkeley Neural Parser \cite[benepar]{kitaev2018constituency} to obtain constituent trees for each sentence and recursively extract noun phrases.
To handle coordinated expressions such as ``broad and high nasal bridge'', which corresponds to both ``broad nasal bridge'' and ``high nasal bridge'' concepts, we introduce a coordinated phrase decomposition mechanism. Using spaCy's dependency parser \citep{honnibal2020spacy}, we identify conjuncts for each coordinated token and then determine its longest constituent not overlapping with its conjuncts. Higher-level noun phrases covering multiple such constituents serve as templates to produce decomposed entities, where only one coordinated token's constituent is used at once. This approach, compared with PhenoBERT which only handles A+B coordination, is capable of generalizing to more complex coordinated structures.
Finally, we integrate the abbreviation detector from scispaCy \citep{neumann-etal-2019-scispacy} to identify abbreviations and replace them by corresponding long forms, in order to reduce mismatch between abbreviations and unrelated identical concepts.

% \subsection{Candidate concept retrieval}
% \label{sec:ccr}

\begin{definition}[Candidate concept retrieval] Given an ontology $O{=}(C, I)$, an entity $e{=}T[i\mathord{:}j]$ from text $T$, and a maximum number of candidates $k$, the task is to find $f_{\text{CCR}}$ satisfying
\begin{equation}
    \begin{aligned}
        f_{\text{CCR}}(O, e, k) &= C_\text{cand}\subset C \\
        \text{s.t.}\quad (|C_\text{cand}|\leq k,\; e \sim \exists c\in C_\text{cand}&) \vee (C_\text{cand}=\emptyset,\; e\not\sim \forall c\in C)
    \end{aligned}
\end{equation}
% \begin{equation}
%     \begin{aligned}
%         f_{\text{CCR}}(O, e, k) = \ & C_\text{cand}\subset C \\
%         \text{s.t. } \quad & (|C_\text{cand}|\leq k, e \sim \exists c\in C_\text{cand}) \vee (C_\text{cand}=\emptyset, e\not\sim \forall c\in C)
%     \end{aligned}
% \end{equation}
% \begin{equation}
%     \begin{aligned}
%         f_{\text{CCR}}(O, e, k) =\ & C_\text{cand}\subset C\\
%         \text{s.t.}\quad\ & |C_\text{cand}|\leq k,\quad e\sim \exists c\in C_\text{cand}\\
%         \vee\quad\ & C_\text{cand}=\emptyset,\quad e\not\sim \forall c\in C
%     \end{aligned}
% \end{equation}
\label{def:cdr}
which retrieves a small set of ontology concepts that potentially match each extracted entity.
\end{definition}

To retrieve candidate concepts for each extracted entity $e$ from the input text, we adopt an embedding-based retrieval strategy with SapBERT, an embedding model trained to align UMLS concept names and synonyms. All ontology concepts and their synonyms are embedded in advance to construct a dense vector index, and then each extracted entity is embedded and compared against the index based on cosine similarity.
If the highest similarity score exceeds a high-confidence threshold $\tau_1$, the entity is directly linked to the most similar concept. Otherwise, if the score falls within a relaxed interval $[\tau_2, \tau_1)$, we retrieve the top-$k$ most similar concepts to form the candidate set $C_\text{cand}$ for downstream LLM-based entity linking. This hierarchical approach ensures high precision for unambiguous matches while preserving recall by delegating ambiguous cases to the more capable LLM.

% \subsection{Entity linking}
% \label{sec:el}

\begin{definition}[Entity linking] Given an ontology $O{=}(C, I)$, an entity $e{=}T[i\mathord{:}j]$ from text $T$, and candidate concepts $C_\text{cand}$, the task is to find $f_{\text{EL}}$ satisfying
\begin{equation}
    \begin{aligned}
        f_{\text{EL}}(O, e, C_\text{cand}) = c\in C_\text{cand} \sim e
    \end{aligned}
\end{equation}
\label{def:el}
which selects the most semantically similar concept from the retrieved candidate concepts.
\end{definition}

Inspired by the prompt-based method REAL, we perform entity linking by prompting an LLM. For each extracted entity $e$ and its retrieved candidate concept set $C_\text{cand}$, we construct a structured prompt that includes the entity name along with the list of candidate concepts. Each concept is represented by its ID, name, definition, synonyms, and cross-referenced UMLS synonyms. The output contains a concept ID or ``None'' with a confidence level, and only predictions with ``HIGH'' confidence are retained to reduce false positives. Specifically, the following prompt template is used.

\begin{quote}
\textbf{System prompt:} \\
As an expert clinician, your task is to accurately link the entity using the concepts listed below. Accuracy is paramount. If the entity does not precisely refer to any of the concepts listed below, please return ``None''; otherwise, return the corresponding concept ID in the following format: \\
answer:$<$concept ID or None$>$ \\
confidence:$<$one of HIGH, LOW, MEDIUM$>$ \\[0.5em]
% \\
Below are the concepts: \\
\{\textit{candidate concepts with ID, name, definition, synonyms, and UMLS synonyms.}\} \\[0.5em]
% \\
\textbf{User prompt:} \\
\lbrack Entity to link\rbrack \\
label: \{\textit{entity string}\}
\end{quote}

To facilitate the LLM-based entity linker to distinguish between highly similar concepts within the ontology, we introduce a self-supervised fine-tuning strategy. We first generate challenging training examples by leveraging the internal structure of the ontology and the retrieval capabilities of SapBERT. For each ontology concept, we perform synonym cleaning by lemmatizing all associated synonyms and merging those that become identical. For each unique lemmatized synonym $s$ belonging to a ground-truth concept $c_{gt}$, we use SapBERT to retrieve the top-$r$ non-repetitive candidate concepts. We then construct positive and negative training examples as follows: To create a positive candidate set $C_\text{cand}^+$, we check if $c_{gt}$ is present in the retrieved list. If it is, the list is preserved. If not, we replace the last concept in the list with $c_{gt}$ to ensure the ground truth is present among difficult distractors. Note that $s$ should be removed from the synonym list of $c_{gt}$ to avoid leakage. To create a negative candidate set $C_\text{cand}^-$, we ensure $c_{gt}$ is absent. If $c_{gt}$ is in the list, we remove it and append the $(r{+}1)$-th retrieved concept to the end. If not, we retain the list as is. This procedure yields two candidate sets of length $r$ for each synonym, representing the most difficult cases for SapBERT to resolve. We transform these examples into training instances by applying the prompt template above with two modifications: (1) we remove the ``confidence'' and ``UMLS synonyms'' requirements from the system prompt to simplify the learning objective, and (2) we set the desired output to the ground-truth concept ID for positive examples and ``None'' for negative examples. By iterating through all concept synonyms, we generate a total of 42,110 training instances. We fine-tune the entity linker using QLoRA \citep{dettmers2023qlora} for one epoch with $r=3$ candidates to achieve efficient adaptation, and name the resulting model $\text{AutoPCR}_\text{FT}$ from now on.

% \subsection{Post-processing}

For post-processing, AutoPCR may produce overlapping entity spans linked to the same or different concepts. When linked to different concepts, all spans are retained. When linked to the same concept, the span with the highest similarity score is prioritized over those linked by the LLM. AutoPCR also provides an option to keep only the longest spans, similar to other neural methods, for scenarios where nested spans are not required.

\section{Experiments}

We conduct extensive experiments to answer three research questions regarding our AutoPCR and its fine-tuned version AutoPCR$_\text{FT}$: (\textbf{RQ-1}) How does AutoPCR perform against other baselines on various datasets? (\textbf{RQ-2}) How much does each module of AutoPCR contribute to its performance? (\textbf{RQ-3}) Can AutoPCR generalize effectively and efficiently to a different ontology?
Benchmarks, baselines, and implementation details will be introduced first, followed by experiment results and analyses answering the questions.

\paragraph{Benchmarks and baselines}
\label{sec:benchmark}

To evaluate the performance of AutoPCR, we conduct experiments on four widely-used datasets: 
\textbf{BIOC-GS}, the development set from BioCreative VIII Track 3 \citep{weissenbacher2023phenoid}, which consists of 382 clinical observation records from dysmorphology physical examinations (\added{avg.} 8.5 words, \added{e.g.}, ``ABDOMEN: Small umbilical hernia. Mild distention. Soft.'').
% It includes 607 phenotype mentions covering 315 unique HPO concepts. 
\textbf{GSC-2024}, a refined GSC+ \citep{lobo2017identifying} dataset given by FastHPOCR, which comprises 228 PubMed abstracts (\added{avg.} 150 words). We use 22 of them for development \added{and the rest for evaluation}, following PhenoTagger's data split.
% The test split contains 2,034 phenotype mentions linked to 451 unique HPO concepts, with an average of 150 words.
\textbf{ID-68}, 68 real clinical notes \citep{anazi2017expanding} from families with intellectual disabilities (\added{avg.} 157 words). It is manually annotated by PhenoBERT in the same way as GSC+. \added{No splits are provided so we use all of them for evaluation.}
% It includes 857 phenotype mentions covering 433 unique HPO concepts, with an average length of 157 words.
\textbf{NCBI} \citep{dougan2014ncbi}, \added{of which we use the test split} of 100 PubMed abstracts (\added{avg.} 205 words).
% It includes 960 phenotype mentions covering 198 unique MEDIC concepts.
\added{The ontologies used are HPO release 2024-02-08 under the root ``phenotypic abnormality'' (17,542 concepts) for BIOC-GS, GSC-2024, and ID-68, and MEDIC \citep{davis2025comparative}, a curated disease ontology integrating MeSH \citep{lipscomb2000medical} and OMIM \citep{amberger2019omim}, under the root ``diseases'' (13,315 concepts) for NCBI.}
% Detailed statistics of the datasets are provided in Appendix~\ref{sec:stat}.
These datasets involve both free-form and standardized concept mentions, covering a wide range of use cases.
% (i) GSC+ \citep{lobo2017identifying}, which comprises 228 scientific abstracts from PubMed. We followed \citeposs{luo2021phenotagger}'s data split, using 22 abstracts for development. The dataset contains 1,949 phenotype mentions linked to 405 unique HPO concepts, and has an average of approximately 150 words per abstract.

We follow the evaluation pipeline of PhenoTagger and PhenoTagger++, except that each predicted and gold entity is counted at most once to ensure one-to-one matching. Mention- and document-level precision (P\%), recall (R\%), and F1\% are used as evaluation metrics. For mention-level evaluation, a predicted entity is considered correct only if there exists a ground-truth entity with the same linked concept and overlapping offsets. For document-level evaluation, linked concepts are treated as a set per document, and metrics are computed by comparing these sets, followed by micro-averaging over all documents. Average results over three runs are reported for all experiments.
% \texttt{HP:0000118}, \texttt{MESH:C}

We compare AutoPCR with competitive baselines from three categories: (1) dictionary-based methods, including NCBO, OBO, ClinPhen, and FastHPOCR; (2) neural methods, including NCR, PhenoTagger, PhenoBERT, and PhenoTagger++; and (3) prompt-based methods, including \added{Vanilla, using the best-performing zero-shot template (``Prompt~4'') from \citet{groza2024evaluation},} and REAL.
All baselines are tested with their recommended parameters. Prompt-based methods employ GPT-4o-mini as the LLM backend.

\paragraph{Implementation details}
\label{sec:implementation}

% We adopt different entity extraction strategies for each dataset, as described in Section~\ref{sec:eg}. For BIOC-GS, we use the rule-based strategy from PhenoTagger, which suits shorter, free-form clinical text. For GSC-2024, ID-68, and NCBI, we apply the neural tagging approach from PhenoBERT, as these datasets consist of longer, standardized content.
For entity extraction, we use Stanza packages ``ner-i2b2'' and ``mimic'', spaCy model ``en\_core\_web\_trf'', benepar model ``benepar\_en3\_large'', and scispaCy model ``en\_core\_sci\_scibert''. We recover abbreviations on ID-68, as they are counted separately. For candidate concept retrieval, we set $\tau_1 {=} 0.95{\in} \{0.9, 0.925, 0.95, 0.975, 1\}$ and $\tau_2 {=} 0.85{\in} \{0.8, 0.825, 0.85, 0.875, 0.9\}$ to distinguish high- and low-confidence matches. For similarity scores in $[\tau_2, \tau_1)$, we retain up to $k {=} 5{\in} \{3, 5, 7, 9\}$ candidates. These hyperparameters are tuned on the GSC-2024 development set. For entity linking, we use Qwen3-Next-80B-A3B-Instruct as the frozen LLM backend and Qwen3-30B-A3B-Instruct-2507 as the base for fine-tuning \citep{yang2025qwen3}. Temperature is fixed at zero for stability.

% (16 cores, 3.1GHz)
% All experiments are conducted on a Linux server equipped with an Intel Xeon Gold 6242R CPU (16 cores at 3.1GHz), 128 GB of memory, and a single NVIDIA Tesla V100 GPU with 32 GB of VRAM.

\paragraph{Main results}
\label{sec:main}

\begin{table*}[!t]
    \caption{Mention-level results on three datasets. AutoPCR is the highest in average F1 and F1 ranking.}
    \label{tab:men}
    \begin{tabular*}{\textwidth}{@{\extracolsep{\fill}}llccccccccccc@{\extracolsep{\fill}}}
\hline
\multicolumn{2}{c}{\multirow{2}{*}{Method}} & \multicolumn{3}{c}{BIOC-GS} & \multicolumn{3}{c}{GSC-2024} & \multicolumn{3}{c}{ID-68} & \multirow{2}{*}{Avg. F1} & \multirow{2}{*}{Avg. Rk.} \\ \cline{3-11}
\multicolumn{2}{c}{} & P & R & F1 (Rk.) & P & R & F1 (Rk.) & P & R & F1 (Rk.) &  &  \\ \hline
\multirow{4}{*}{\begin{tabular}[c]{@{}l@{}}Dictionary-\\ Based\end{tabular}} & NCBO & \underline{82.20} & 45.63 & 58.69 (8) & \textbf{95.71} & 51.57 & 67.03 (7) & 86.27 & 65.23 & 74.29 (7) & 66.67 & 7.33 \\
 & OBO & 71.94 & 42.67 & 53.57 (10) & 85.89 & 52.66 & 65.29 (8) & 81.41 & 60.79 & 69.61 (9) & 62.82 & 9.00 \\
 & ClinPhen & 64.64 & 50.91 & 56.96 (9) & 84.10 & 38.74 & 53.05 (10) & 72.92 & 60.33 & 66.03 (10) & 58.68 & 9.67 \\
 & FastHPOCR & 66.73 & 59.47 & 62.89 (5) & \underline{88.43} & \textbf{79.25} & \textbf{83.59} (1) & \underline{87.23} & 71.76 & 78.75 (3) & 75.08 & \underline{3.00} \\ \hline
\multirow{4}{*}{Neural} & NCR & 64.04 & 57.50 & 60.59 (7) & 68.54 & 74.88 & 71.57 (6) & 78.64 & \underline{78.18} & 78.41 (4) & 70.19 & 5.67 \\
 & PhenoTagger & 71.12 & \underline{65.73} & \underline{68.32} (2) & 86.16 & 78.07 & 81.92 (4) & 84.40 & 72.58 & 78.04 (5) & \underline{76.09} & 3.67 \\
 & PhenoBERT & 74.77 & 53.71 & 62.51 (6) & 85.90 & 74.88 & 80.01 (5) & \textbf{93.36} & \textbf{78.76} & \textbf{85.44} (1) & 75.99 & 4.00 \\
 & PhenoTagger++ & 69.16 & 63.92 & 66.44 (4) & 87.71 & 78.22 & 82.69 (3) & 79.90 & 72.81 & 76.19 (6) & 75.11 & 4.33 \\ \hline
\multirow{4}{*}{\begin{tabular}[c]{@{}l@{}}Prompt-\\ Based\end{tabular}} & Vanilla & 2.91 & 2.31 & 2.57 (11) & 15.31 & 7.57 & 10.13 (11) & 18.29 & 11.20 & 13.89 (11) & 8.87 & 11.00 \\
 & REAL & 75.68 & 59.97 & 66.91 (3) & 76.19 & 47.35 & 58.40 (9) & 76.33 & 65.46 & 70.48 (8) & 65.26 & 6.67 \\
 & $\text{AutoPCR}$ & \textbf{84.25} & \textbf{67.27} & \textbf{74.81} (1) & 87.13 & \underline{79.19} & \underline{82.97} (2) & 83.68 & 76.59 & \underline{79.98} (2) & \textbf{79.25} & \textbf{1.67} \\
 \cdashline{2-13}
 & $\text{AutoPCR}_\text{FT}$ & 83.44 & 65.02 & 73.09 & 89.08 & 77.94 & 83.14 & 87.84 & 77.01 & 82.07 & 79.43 & --- \\ \hline
\end{tabular*}
\end{table*}

% \begin{table*}[!t]
%     \caption{Document-level results on three datasets. AutoPCR is the highest in average F1 and F1 ranking.}
%     \label{tab:doc}
%     \input{tables/doc}
% \end{table*}

\begin{table}[!t]
    \caption{Mention-level recall on coordinated entities from the BIOC-GS, GSC-2024, and ID-68 datasets. AutoPCR achieves the highest recall.}
    \label{tab:coord}
    \begin{tabular*}{\columnwidth}{@{\extracolsep\fill}llc@{\extracolsep\fill}}
\hline
\multicolumn{2}{c}{Method} & Coordinated Entities \\ \hline
\multirow{4}{*}{\begin{tabular}[c]{@{}l@{}}Dictionary-\\ Based\end{tabular}} & NCBO & 5.98 \\
 & OBO & 26.63 \\
 & ClinPhen & 13.04 \\
 & FastHPOCR & 9.24 \\ \hline
\multirow{4}{*}{Neural} & NCR & 15.76 \\
 & PhenoTagger & 19.57 \\
 & PhenoBERT & 21.74 \\
 & PhenoTagger++ & 25.00 \\ \hline
\multirow{5}{*}{\begin{tabular}[c]{@{}l@{}}Prompt-\\ Based\end{tabular}} & Vanilla & 1.09 \\
 & REAL & \underline{38.59} \\
 & $\text{AutoPCR}$ (w/o Coord.) & 22.28 \\
 & $\text{AutoPCR}$ & \textbf{45.11} \\
 % \cdashline{2-3}
 % & $\text{AutoPCR}_\text{FT}$ (w/o Coord.) & X \\
 % & $\text{AutoPCR}_\text{FT}$ & 38.41 \\
 \hline
\end{tabular*}

% \begin{tabular*}{\textwidth}{@{\extracolsep{\fill}}l cccc cccc cccccc}
% \hline
% \multicolumn{4}{c}{\textbf{Dictionary-Based}} & \multicolumn{4}{c}{\textbf{Neural}} & \multicolumn{6}{c}{\textbf{Prompt-Based}} \\
% \textbf{Method} & NCBO & OBO & ClinPhen & FastHP & NCR & PT & PB & PT++ & Van. & REAL & \begin{tabular}[c]{@{}c@{}}Auto\\ (w/o)\end{tabular} & Auto & \begin{tabular}[c]{@{}c@{}}Auto$_{FT}$\\ (w/o)\end{tabular} & Auto$_{FT}$ \\
% \hline
% \textbf{Coordinated Entities} & 5.98 & 26.63 & 13.04 & 9.24 & 15.76 & 19.57 & 21.74 & 25.00 & 1.09 & 38.59 & 23.37 & \textbf{45.11} & 23.37 & \textbf{45.11} \\
% \hline
% \end{tabular*}
    % AutoPCR achieves the highest recall., highlighting the effect of coordinated phrase decomposition.}
\end{table}

The mention- and document-level results across three datasets grounded on HPO are shown in Tables~\ref{tab:men} and \added{Supplementary Table 1, and visualized in \added{Supplementary Figures 1 and 2},} respectively.
The two evaluation levels yield identical method rankings.
% indicating consistent comparative performance
AutoPCR ranks first on BIOC-GS and second on both GSC-2024 and ID-68, achieving the highest average F1 and the best overall rank.
It also exhibits the most stable performance, demonstrating strong robustness across datasets.
FastHPOCR achieves the strongest performance among dictionary-based methods, with the second-best average rank.
Nevertheless, its performance drops notably on BIOC-GS, indicating reduced robustness on noisier clinical narratives.
Traditional alternatives such as NCBO, OBO, and ClinPhen maintain high precision but suffer from low recall, resulting in overall lower F1.
Neural methods PhenoTagger and PhenoBERT rank third and fourth overall.
PhenoTagger performs better on BIOC-GS but worse on ID-68, whereas PhenoBERT shows the opposite trend.
Both methods exhibit higher variability compared with AutoPCR.
The earlier NCR is outperformed by its successor PhenoBERT, as expected, while PhenoTagger++, intended as a refinement of PhenoTagger, shows only a slight improvement on GSC-2024 but degraded performance on BIOC-GS and ID-68, leading to an overall decline and suggesting possible overfitting. 
Vanilla prompting performs poorly compared with all other baselines.
In contrast, the RAG-based REAL performs strongly on BIOC-GS but ranks near the bottom on the others, highlighting the limitations of its generalist model design when applied to long, standardized text.
These results comprehensively answer (\textbf{RQ-1}), demonstrating the \added{strong} and robust performance of AutoPCR across datasets.
We further analyze these outcomes in detail, explaining how the design of AutoPCR contributes to its performance under different dataset characteristics.

% where it shows relatively lower F1 scores compared to neural and prompt-based methods.

% While they performs strongly on GSC-2024 and ID-68, it underperforms on BIOC-GS, likely due to the dataset’s unstructured clinical notes, which are less suited to neural models trained on more formal biomedical text.
% where it shows relatively lower F1 scores compared to neural and prompt-based methods.

% When comparing different method categories, dictionary-based methods tend to achieve high precision but suffer from low recall, resulting in lower F1 scores on unstructured datasets. Neural methods show competitive performance overall, but exhibit more variance across datasets. In contrast, prompt-based methods (especially AutoPCR) offer a better balance, with both high recall and strong cross-domain adaptability.

On BIOC-GS consisting of noisy clinical notes, we observe a clear trend across method categories: prompt-based methods outperform neural methods, which in turn surpass dictionary-based ones, consistent with prior findings \citep{groza2024evaluation, qi2024improved, shlyk2024real}. This pattern reflects their respective modeling capabilities. Prompt-based methods use LLMs with strong language understanding abilities for entity linking, enabling them to better handle free-form and lexically diverse clinical text. Neural methods, typically fine-tuned on structured, ontology-aligned corpora, struggle to generalize to such noisy input. Dictionary-based methods perform worst, as they lack semantic understanding and rely solely on exact or approximate string matching. 
% Within the prompt-based category, for the same reason, REAL performs better than AutoPCR due to its use of LLM-generated definition-similarity-based concept retrieval, which relies less on the surface forms of the entities.

On GSC-2024 and ID-68, which feature more standardized text with less surface-level variation in entities, AutoPCR remains highly competitive, even though the advantage of language understanding becomes less pronounced. The top-performers on these two datasets, FastHPOCR and PhenoBERT, \added{do} not bring the advantage to the other; by contrast, AutoPCR consistently ranks behind them in the second place. This may be possibly because both methods partially annotated the datasets themselves, introducing inductive bias in their favor.
% Specifically, ID-68, as annotated by PhenoBERT, contains mentions that are not exactly matched by ontology concepts but instead linked to hypernyms (e.g., ``thinning of the body of the corpus callosum'' linked to HP:0002079 ``Hypoplasia of the corpus callosum''). While this criterion may be helpful in certain use cases, it diverges from the standard task definition shared by our study and other competitive baselines. This inconsistency creates an ambiguous annotation standard: if every term covered by the ontology's hierarchy is a potential candidate, the task essentially shifts into a classical hierarchical classification problem. Conversely, if only a subset is chosen, the labels become heavily influenced by the idiosyncratic styles of the annotators, making the problem more suitable for supervised learning rather than the zero-shot setting we address. 
This is also reflected in the experimental results, where PhenoBERT achieves an outlier precision on ID-68, hinting at a bias in the dataset's construction \added{(more in the Discussion section)}. Nevertheless, AutoPCR still secures the second best, outperforming the next best baseline by approximately 3\% in F1. Similarly, AutoPCR maintains a clear advantage over all other baselines except FastHPOCR on GSC-2024. This consistent lead demonstrates \added{that AutoPCR remains robust even under deviating annotation criteria.}
Unlike on BIOC-GS, AutoPCR here significantly outperforms REAL, which can be credited to two novel designs: (1) a comprehensive entity extraction approach that improves recall on long, standardized text compared to LLM-based extraction and (2) a candidate retrieval module that effectively models domain-specific semantic similarity to filter noisy entities while prioritizing high-quality candidates.

The findings for AutoPCR are largely consistent with those of $\text{AutoPCR}_{\text{FT}}$, as they have a similar average F1 and share identical rankings if ranked separately. However, by fine-tuning on HPO with mostly standardized synonyms, $\text{AutoPCR}_{\text{FT}}$ trades off some of its ability to handle free-form text for enhanced precision in linking standardized entities. This shift is evidenced by a slight decrease in F1 on BIOC-GS, contrasted with a precision gain on GSC-2024 and ID-68. Specifically, on ID-68, both recall and precision improve simultaneously, leading to an F1 increase of over 2.0. These results validate our self-supervised fine-tuning strategy and demonstrate the effectiveness of ontology-specific adaptation when the input text style is aligned with the ontology's.

\begin{table*}[!t]
    \caption{Ablation study of AutoPCR on syntactic entity extraction, HPO/UMLS knowledge, and entity linking. Removing any component overall degrades performance. mF1: mention-level F1; dF1: document-level F1.}
    \begin{tabular*}{\textwidth}{@{\extracolsep{\fill}}lcccccccccccc@{\extracolsep{\fill}}}
\hline
\multicolumn{1}{c}{\multirow{2}{*}{}} & \multirow{2}{*}{\begin{tabular}[c]{@{}c@{}}Syntactic\\ Entities\end{tabular}} & \multirow{2}{*}{\begin{tabular}[c]{@{}c@{}}HPO\\ Knowledge\end{tabular}} & \multirow{2}{*}{\begin{tabular}[c]{@{}c@{}}UMLS\\ Knowledge\end{tabular}} & \multirow{2}{*}{\begin{tabular}[c]{@{}c@{}}Entity\\ Linking\end{tabular}} & \multicolumn{2}{c}{BIOC-GS} & \multicolumn{2}{c}{GSC-2024} & \multicolumn{2}{c}{ID-68} & \multirow{2}{*}{Avg. mF1} & \multirow{2}{*}{Avg. dF1} \\ \cline{6-11}
\multicolumn{1}{c}{} &  &  &  &  & mF1 & dF1 & mF1 & dF1 & mF1 & dF1 &  &  \\ \hline
Variant 1 &  & $\checkmark$ & $\checkmark$ & $\checkmark$ & 67.58 & 67.84 & 82.18 & 82.94 & 79.38 & 79.60 & 76.38 & 76.79 \\
Variant 2 & $\checkmark$ &  & $\checkmark$ & $\checkmark$ & \underline{74.10} & \underline{74.33} & \underline{82.61} & \underline{83.22} & 78.77 & 78.99 & \underline{78.49} & \underline{78.84} \\
\added{Variant 3a} & \added{$\checkmark$} &  &  & \added{$\checkmark$} & \added{60.49} & \added{60.85} & \added{57.05} & \added{59.09} & \added{68.58} & \added{72.67} & \added{62.04} & \added{64.20} \\
\added{Variant 3b} & \added{$\checkmark$} &  &  & \added{$\checkmark$} & \added{73.02} & \added{73.31} & \added{77.77} & \added{79.89} & \added{77.72} & \added{77.77} & \added{76.17} & \added{76.99} \\
Variant 4 & $\checkmark$ & $\checkmark$ & $\checkmark$ &  & 69.90 & 70.19 & 81.43 & 81.37 & \textbf{80.80} & \textbf{80.93} & 77.38 & 77.49 \\
$\text{AutoPCR}$ & $\checkmark$ & $\checkmark$ & $\checkmark$ & $\checkmark$ & \textbf{74.81} & \textbf{75.04} & \textbf{82.97} & \textbf{83.48} & \underline{79.98} & \underline{80.34} & \textbf{79.25} & \textbf{79.62} \\ \hline
\end{tabular*}
    \label{tab:ablation}
\end{table*}

To further evaluate each method's ability to handle coordinated phrases, we curate 184 gold-standard entities from the BIOC-GS, GSC-2024, and ID-68 datasets that contain conjunctions such as ``and'', ``or'', ``/'', ``and/or'', or commas. These entities represent cases where a single phrase encodes multiple coordinated concepts, providing a focused benchmark for testing coordinated phrase decomposition. As shown in Table~\ref{tab:coord}, AutoPCR achieves the highest recall, while its variant without coordinated phrase decomposition performs comparably to other baselines. This confirms the effectiveness of our decomposition mechanism in correctly decomposing coordinated entities into separate concepts. REAL ranks second, as the LLM-based entity extraction can inherently split coordinated mentions, but it still falls short of AutoPCR and generates substantially more output tokens, leading to higher computational and monetary costs. Among neural methods, PhenoBERT, despite claiming to handle coordination like ``A+B'', does not show noticeable improvement over its peers. Within dictionary-based methods, OBO stands out, which is expected since it heuristically concatenates noncontiguous subsequences into entities, capturing simple coordinated cases. To summarize, a representative example may be the phrase ``hearing loss of conductive, sensorineural, or mixed type'', which requires decomposition into three distinct concepts: ``hearing loss of \{conductive, sensorineural, mixed\} type''. Only AutoPCR can achieve this.

\paragraph{Ablation study}
\label{sec:ablation}

% \begin{itemize}[leftmargin=*, itemsep=0.0em]

To assess the contribution of each module in AutoPCR and answer (\textbf{RQ-2}), we conduct ablation experiments shown in Table~\ref{tab:ablation}. Specifically, we evaluate the following four variants of AutoPCR. 
\textbf{Variant 1} removes syntactic entity extraction and relies solely on bioNER for entity extraction. It assesses whether syntactic entities contribute beyond direct BioNER segmentation. 
\textbf{Variant 2} disables HPO knowledge and retrains SapBERT (initialized from PubMedBERT) on UMLS excluding all HPO-related concepts. It simulates a scenario where the target ontology is newly introduced and unseen during model training. 
\added{\textbf{Variant 3a}} examines the necessity of domain-specific alignment in SapBERT by replacing it with PubMedBERT for embedding, without any fine-tuning on UMLS.
\added{\textbf{Variant 3b} replaces SapBERT with \texttt{text-embedding-ada-002}, the general-purpose embedding model used by REAL, to isolate AutoPCR's architectural contribution from its retrieval model choice.}
\added{Implementation details of both variants are in Supplementary Section~1.}
\textbf{Variant 4} evaluates the importance of entity linking by removing the linking module and \added{relying} solely on candidate retrieval for concept prediction. It makes prediction only if the top candidate concept exceeds the similarity threshold $\tau_1$.

The full AutoPCR model overall outperforms all ablated variants across datasets and metrics, confirming the importance of each component.
Variant 1 shows a clear performance decline on BIOC-GS, confirming that BioNER alone is insufficient for handling free-form text and that syntactic entity extraction is essential for improving recall in such cases.
\added{Variant 3a} exhibits the most severe performance drop, demonstrating that domain-specific alignment in embedding-based retrieval is critical for accurate concept grounding. 
% A further experiment replacing SapBERT with REAL's embedding model corroborates this finding and demonstrates AutoPCR's architectural superiority over REAL as an RAG-based framework (Supplementary Section~1).
\added{Variant 3b, trailing AutoPCR by approximately 2 F1 points on average, corroborates this finding, while still exceeding REAL by nearly 10 F1 points, highlighting the strength of AutoPCR as a framework over REAL independent of the retrieval model.}
%demonstrates that even when AutoPCR is equipped with the same general-purpose embedding model as REAL, its average performance still exceeds REAL by nearly 10 F1 points, highlighting the strength of AutoPCR as a framework independent of the retrieval model. At the same time, Variant 3b trails AutoPCR by approximately 2 F1 points on average, validating the choice of domain-specific embedding model SapBERT.}
Variant 4 performs reasonably well on standardized abstracts such as GSC-2024 and ID-68, suggesting that surface-form matching alone may suffice for low-ambiguity datasets. However, it underperforms markedly on BIOC-GS, which contains noisier and more ambiguous clinical narratives, underscoring the value of LLM-based reasoning for fine-grained semantic disambiguation.
Notably, Variant 2 achieves the strongest results among all variants and remains comparable to the full AutoPCR model even without HPO-specific knowledge, indicating that AutoPCR can perform inductive inference on entirely new or rapidly evolving \added{biomedical} ontologies, such as HPO.

% where surface forms offer strong signals
% Given that HPO is updated monthly, it is impractical to retrain neural models frequently. In contrast, AutoPCR’s modular architecture—with plug-and-play retrieval and linking—enables continual adaptation without requiring additional supervision or re-training.

% This highlights the essential role of ontology-aware pretraining in enabling effective retrieval.

% These results collectively affirm that ontology-guided retrieval and LLM-based disambiguation are not only complementary, but also essential for building flexible, scalable, and ontology-agnostic concept recognition systems.

\paragraph{Generalizability}
\label{sec:generalizability}

\begin{table}[!t]
    \caption{Performance of selected methods on NCBI. AutoPCR obtains the highest accuracy with \added{one of} the lowest deployment \added{times}.}
    \begin{tabular*}{\columnwidth}{@{\extracolsep\fill}lccc@{\extracolsep\fill}}
\hline
\multicolumn{1}{c}{\multirow{2}{*}{Method}} & \multicolumn{3}{c}{NCBI} \\ \cline{2-4} 
\multicolumn{1}{c}{} & mF1 & dF1 & Deployment Time (min) \\ \hline
FastHPOCR & 49.54 & 57.11 & 4.2 \\
NCR & 34.24 & 31.80 & 1680 \\
PhenoTagger & \underline{67.48} & \underline{68.83} & 906 \\
\added{Vanilla} & \added{7.63} & \added{10.77} & \added{0} \\
\added{REAL} & \added{41.76} & \added{52.85} & \added{0.4} \\
$\text{AutoPCR}$ & \textbf{73.15} & \textbf{71.93} & 2.6 \\
\cdashline{1-4}
$\text{AutoPCR}_\text{FT}$ & 72.76 & 70.43 & 2.6 \\ \hline
\end{tabular*}
    \label{tab:medic}
\end{table}

To evaluate the generalizability and deployment efficiency of AutoPCR, we conduct an additional experiment on the NCBI dataset using the MEDIC ontology, as shown in Table~\ref{tab:medic}.
% Table~\ref{tab:medic} reports both mention- and document-level results, along with the time required to deploy each method. 
We define \emph{deployment time} as the total time needed to prepare a method for inference on a new ontology, including ontology-specific index construction or model retraining.
% (Appendix~\ref{sec:ncbi_setting})
\added{We compare AutoPCR with representative baselines: FastHPOCR, NCR, PhenoTagger, Vanilla, and REAL.} PhenoTagger++ and PhenoBERT are excluded due to the lack of publicly available pipelines for adapting to a new ontology.

AutoPCR achieves the highest F1 on both mention- and document-level, \added{and is also one of} the most efficient to deploy, requiring only 2.6 minutes to build the index of embedded concepts. 
AutoPCR$_\text{FT}$, though fine-tuned on HPO, maintains most of its generalizability and exhibits minimal overfitting.
In contrast, neural methods such as PhenoTagger and NCR require hours of retraining and still fall short in performance. FastHPOCR deploys more quickly than neural methods but performs substantially worse than both PhenoTagger and AutoPCR, indicating limited generalizability. \added{Vanilla and REAL perform even worse, though with low deployment time as prompt-based methods.} These results answer (\textbf{RQ-3}) and demonstrate that AutoPCR can generalize effectively and efficiently to a new ontology without reconfiguration, \added{making} it well suited for real-world, off-the-shelf use in different biomedical domains.

% Its ability to support rapid deployment makes it well suited for real-world, off-the-shelf use in different biomedical domains.

% To evaluate the generalizability and deployment efficiency of AutoPCR, we conduct an additional experiment on the NCBI Disease Corpus using the MEDIC ontology. Table~\ref{tab:medic} reports both mention-level and document-level results, as well as the time required to deploy each method—defined as the total time needed to prepare a method for inference on a new ontology, including index construction or re-training if applicable. We compare against three strong and representative baselines: FastHPOCR, NCR, and PhenoTagger. PhenoTagger++ and PhenoBERT are excluded due to the absence of publicly available training pipelines for adapting to new ontologies.

% AutoPCR achieves the highest F1 scores on both mention-level (51.47\%) and document-level (66.08\%), while also being the most efficient to deploy, requiring only 2 minutes and 38 seconds to build the ontology index. In contrast, neural methods such as PhenoTagger and NCR require hours of retraining to support a new ontology, yet still underperform relative to AutoPCR. FastHPOCR, though with much faster deployment time than neural methods, lags far behind PhenoTagger and AutoPCR, showing poor generalizability. These results confirm that AutoPCR supports fast and robust adaptation to new ontologies without requiring retraining or prompt re-engineering. Its generalizability combined with low deployment cost makes it suitable for real-time and scalable applications.

\paragraph{Hyperparameter sensitivity}

Lastly, we examine the sensitivity of key hyperparameters in AutoPCR, including the high-confidence threshold $\tau_1$, the low-confidence threshold $\tau_2$, the number of retrieved candidates $k$, and the choice of LLM backends, on the GSC-2024 dataset.
% These analyses assess how each component affects model performance and stability across datasets and configurations.

\begin{figure}[!t]
    \centering
    \includegraphics[width=0.75\columnwidth]{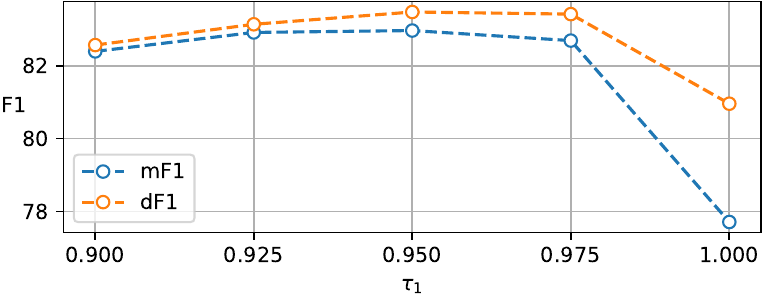}
    \caption{Sensitivity analysis of AutoPCR w.r.t. $\tau_1$ shows robustness.}
    \label{fig:tau1}
\end{figure}

\begin{figure}[!t]
    \centering
    \begin{subfigure}[b]{0.45\columnwidth}
        \centering
        \includegraphics[width=\linewidth]{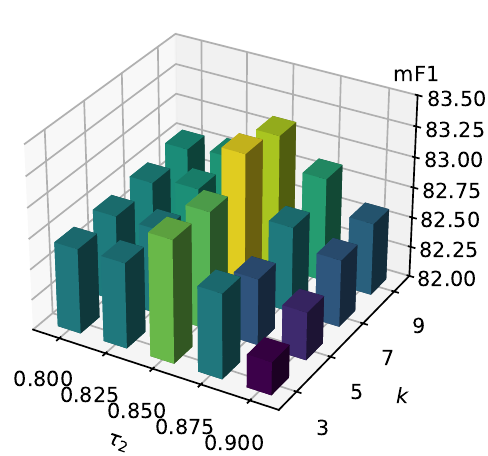}
        % \caption{Mention-level F1}
        \label{fig:tau2_mf1}
    \end{subfigure}
    \hfill
    \begin{subfigure}[b]{0.45\columnwidth}
        \centering
        \includegraphics[width=\linewidth]{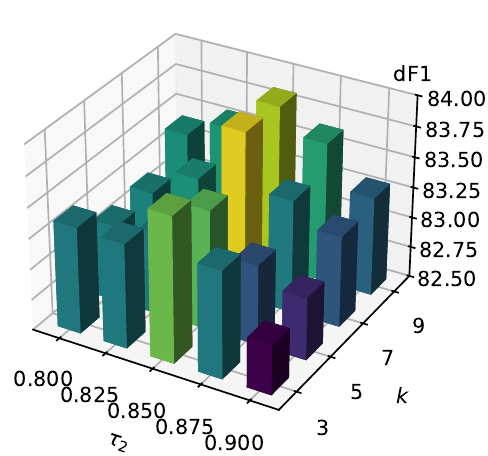}
        % \caption{Dataset-level F1}
        \label{fig:tau2_df1}
    \end{subfigure}
    \caption{Sensitivity analysis of AutoPCR w.r.t. $\tau_2$ and $k$ shows robustness.}
    \label{fig:tau2_k}
\end{figure}

Figures \ref{fig:tau1} and \ref{fig:tau2_k} show the effects of varying $\tau_1$, $\tau_2$, and $k$.
For the high-confidence threshold $\tau_1$, performance improves slightly as $\tau_1$ increases from 0.9 to 0.95, likely because more false positives are filtered by the LLM-based entity linking stage. When $\tau_1$ exceeds 0.95, performance begins to decline and drops sharply at $\tau_1{=}1.0$, where all entities above the low-confidence threshold are forced into the linking stage. This indicates that the LLM can introduce false negatives when surface-form matches are already sufficient, underscoring the need to retain a high-confidence threshold for direct predictions. Overall, AutoPCR is robust to $\tau_1$ across the range of 0.9--0.975, with the default $\tau_1{=}0.95$ achieving the best F1.
When varying $\tau_2$ and $k$, which jointly determine the number and quality of candidate concepts for entity linking, performance is largely stable for $\tau_2{\in}[0.8, 0.875]$, peaking at the default $\tau_2{=}0.85$. Smaller $\tau_2$ introduces excessive noise by retrieving too many irrelevant candidates, whereas larger $\tau_2$ risks missing valid ones, both leading to a degradation in F1. On the other hand, increasing $k$ from 3 to 7 yields marginal performance gains at the expense of increased computational overhead, representing a typical trade-off between accuracy and efficiency.

% $k{=}5$ provides a reasonable compromise between recall and computational efficiency.

\added{Supplementary Table 2} summarizes the performance of AutoPCR across eleven representative non-reasoning LLM backends \citep{grattafiori2024llama, meta2025llama, liu2024deepseek, team2025kimi} under three random seeds. We have the following findings:
(1) The choice of LLM family has only a modest effect on performance. Flagship models from different families, such as Llama-4-Scout, Qwen3-Next-80B, DeepSeek-V3.1, and GPT-4.1, exhibit a performance delta of less than 1.0 F1, demonstrating the robustness of AutoPCR across various LLM backends.
(2) Random seeds have negligible impact. The variance in F1 remains below 0.2 for most backends except Kimi-K2 and Llama-4-Scout, confirming the stability of AutoPCR.
(3) Performance generally scales with model size. Within the Llama family, F1 scores rise steadily from the 3B to the 70B and 109B (Scout) models. Similar upward trends are observed across the GPT family (from nano to mini to GPT-4.1) and the Qwen series (from 30B to 80B). The large-scale DeepSeek-V3.1 (685B) also performs strongly, ranking near the top. These results highlight the scalability of AutoPCR and its ability to benefit from increasingly capable LLMs.
(4) For practical deployment, the open-source Qwen3-Next-80B offers the best overall performance with reasonable inference efficiency, while for proprietary settings, smaller GPT-mini models provide an excellent trade-off between accuracy and cost.

\section{Discussion}

\added{Through error analysis of AutoPCR on ID-68, we identified several annotation issues that likely introduce spurious errors.
First, ID-68, as annotated by PhenoBERT's team, contains mentions linked to hypernyms rather than exact ontology matches. This occurs because PhenoBERT accepts ancestral relationships as correct predictions when an exact match is absent. There are fifteen such cases where PhenoBERT correctly predicts an ancestral relationship that AutoPCR does not (e.g., ``raised choline levels in the white matter'' $\rightarrow$ HP:0002500 \texttt{Abnormal cerebral white matter morphology}).
While this criterion may be useful in certain contexts, it diverges from the standard task definition shared by our study and other competitive baselines, which leads to an ambiguous annotation standard. If every term covered by the ontology’s hierarchy is a potential candidate, the task essentially shifts into a classical hierarchical classification problem. Conversely, if only a subset is chosen, the labels become heavily influenced by annotator-idiosyncratic choices, making the problem more suitable for supervised learning rather than the zero-shot setting we address.
Second, ID-68 was annotated using an outdated version of HPO (2019-09-06), causing several concept labels to differ from those in the 2024-02-08 version adopted by us and most recent methods FastHPOCR and PhenoTagger++. For example, five mentions of ``thin corpus callosum'' are linked to HP:0002079 \texttt{Hypoplasia of the corpus callosum}, whereas HP:0033725 \texttt{Thin corpus callosum} exists as a distinct concept in the newer HPO. Similar issues affect seventeen instances of ``hypotonia'' and three of ``high arched palate''.
Third, several annotations are inconsistent with ID-68's stated convention of preferring longest-span matches. For example, ``poor attention and hyperactivity'' is annotated as separate entities rather than being linked to HP:0007018 \texttt{Attention deficit hyperactivity disorder}, and ``Arnold Chiari malformation type I'' is labeled as ``Arnold Chiari malformation'' instead of HP:0007099 \texttt{Chiari type I malformation}.
These findings highlight the potential for a refined version of ID-68, similar to the transition from GSC+ to GSC-2024, to better support future benchmarking.}

\added{Apart from defective ID-68 benchmarking, AutoPCR has several other limitations as an intelligent system aimed at large-scale deployment. First, the current benchmarks are relatively small and domain-specific. Performance on larger corpora such as radiology and pathology reports remains to be validated. Additionally, AutoPCR does not explicitly handle clinical uncertainty, treating mentions such as ``possible'' or ``rule out'' identically to definitive ones, which may introduce false positives in settings where distinguishing confirmed from suspected phenotypes is critical. Furthermore, its three-stage pipeline introduces inherent sequential dependencies between stages, limiting throughput at scale. System-level optimizations such as pipeline parallelism may be explored to improve efficiency. These directions are left for future work.}

\section{Conclusion}
% TODO: check "Fast Semidifferential-based Submodular Function Optimization"

In this work, we have developed AutoPCR, \added{a prompting-based automated phenotype CR method that decomposes the task into entity extraction, candidate concept retrieval, and LLM-based entity linking. Experiments show that AutoPCR achieves superior average accuracy and robustness across benchmarks, strong inductive capability to handle HPO updates, and efficient generalizability to a new ontology within minutes,} hence the name ``automated''.

%Future work may further improve the adaptability of AutoPCR. Enabling cross-lingual concept recognition helps broader application to non-English biomedical corpora. Leveraging richer contextual signals may improve entity linking in more complex scenarios. Integrating AutoPCR into downstream pipelines such as biomedical knowledge graph construction and phenotype-driven disease diagnosis, in order to assess its utility in real-world biomedical applications.}

\section{Conflicts of interest}
The authors declare that they have no competing interests.

\section{Funding}
% The authors would like to acknowledge the support of NIH awards U24DK138515 and OT2OD038003.
This work is supported by NIH \#U24DK138515 and \#OT2OD038003.

\section{Data availability}

Datasets and ontologies are sourced from \citet[BIOC-GS]{weissenbacher2023phenoid}, \citet[GSC-2024]{groza2024fasthpocr}, \citet[ID-68]{feng2022phenobert}, \citet[NCBI]{luo2021phenotagger}, \citet[HPO]{gargano2024mode}, and \citet[MEDIC]{davis2025comparative}. Access links are available in our code repository.
% BIOC-GS is from the BioCreative VIII Track 3 \citep{weissenbacher2023phenoid}. GSC-2024 is from FastHPOCR \citep{groza2024fasthpocr}. ID-68 is from PhenoBERT \citep{feng2022phenobert}. NCBI is from PhenoTagger \citep{luo2021phenotagger}. HPO is from \citet{gargano2024mode}. MEDIC is from \citet{davis2025comparative}. Access links are available in our code repository.

\section{Author contributions statement}
Y.T., Y.H., and J.L. conceived the methodology. Y.T. and Y.W. conducted the experiments. Y.T., Y.H., Y.W., and X.L. analyzed the results. Y.T. and J.L. wrote and reviewed the manuscript.

%UNCOMMENT THE BELOW TWO LINES IN CASE YOU NEED NUMBERED FORMAT.
% \bibliographystyle{oup-plain}
% \bibliography{reference}

%UNCOMMENT THE BELOW TWO LINES IN CASE YOU NEED AUTHOR YEAR FORMAT.
\bibliographystyle{oup-abbrvnat}
\bibliography{reference}

%% sample for biography with author's image
% \begin{biography}{{\color{black!20}\rule{77pt}{57pt}}}{\author{Yicheng Tao.} This is sample author biography text. The values provided in the optional argument are meant for sample purposes. There is no need to include the width and height of an image in the optional argument for live articles. This is sample author biography text this is sample author biography text this is sample author biography text this is sample author biography text this is sample author biography text this is sample author biography text this is sample author biography text this is sample author biography text.}
% \end{biography}

%% sample for biography without author's image
% \begin{biography}{}{\author{Author Name.} This is sample author biography text this is sample author biography text this is sample author biography text this is sample author biography text this is sample author biography text this is sample author biography text this is sample author biography text this is sample author biography text.}
% \end{biography}

%%%%%%%%%%%%%%

\begin{appendices}

\end{appendices}

\end{document}